\documentclass[letterpaper, 10 pt, conference]{IEEEtran}
\IEEEoverridecommandlockouts
\usepackage{fancyhdr}
\usepackage[utf8]{inputenc}
\usepackage[T1]{fontenc}
\usepackage{cite}
\makeatletter
\let\NAT@parse\undefined
\makeatother
\usepackage{graphics}
\usepackage{epsfig}
\usepackage{mathptmx}
\usepackage{amsmath}
\usepackage{amssymb}
\usepackage{makecell}
\usepackage{multirow}
\usepackage{graphicx}
\usepackage{subfigure}
\usepackage[normalem]{ulem}
\useunder{\uline}{\ul}{}
\usepackage{algorithm}
\usepackage{algorithmic}
\usepackage{geometry}
\title{\LARGE \bf Detecting Lane and Road Markings at A Distance with Perspective Transformer Layers}
\fancypagestyle{FirstPage}{
\fancyhead[L]{
  \copyright~2020 IEEE. Personal use of this material is permitted. Permission from IEEE must be obtained for all other uses, in any current or future media, including reprinting/republishing this material for advertising or promotional purpose, creating new collective works, for resale or redistribution to servers or lists, or reuse of any copyrighted component of this work.}
\fancyfoot[]{}
}

\author{
    Zhuoping Yu$^{1, 2}$, Xiaozhou Ren$^{1, 2}$, Yuyao Huang$^{1, 2}$, Wei Tian$^{1, 2}$, Junqiao Zhao$^{2, 3, *}$
    \thanks{$^{1}$ School of Automotive Studies, Tongji University, Shanghai, China}
    \thanks{$^{2}$ Institute of Intelligent Vehicles, Tongji University, Shanghai, China}
	\thanks{$^{3}$ Department of Computer Science and Technology, Tongji University, Shanghai, China}
    \thanks{$^{*}$ Corresponding author. E-mail: zhaojunqiao@tongji.edu.cn}
}

\usepackage{xspace}
\makeatletter
\DeclareRobustCommand\onedot{\futurelet\@let@token\@onedot}
\def\@onedot{\ifx\@let@token.\else.\null\fi\xspace}

\def\eg{\emph{e.g}\onedot} 
\def\ie{\emph{i.e}\onedot} 
 
\def\etc{\emph{etc}\onedot} 

\def\etal{et al\onedot}
\makeatother

\begin{document}

\maketitle

\thispagestyle{FirstPage}
\pagestyle{empty}
\begin{abstract}
    Accurate detection of lane and road markings is a task of great importance for intelligent vehicles. In existing approaches, the detection accuracy often degrades with the increasing distance. This is due to the fact that distant lane and road markings occupy a small number of pixels in the image, and scales of lane and road markings are \textit{inconsistent} at various distances and perspectives. The Inverse Perspective Mapping (IPM) can be used to eliminate the perspective distortion, but the inherent interpolation can lead to artifacts especially around distant lane and road markings and thus has a negative impact on the accuracy of lane marking detection and segmentation. To solve this problem, we adopt the Encoder-Decoder architecture in Fully Convolutional Networks and leverage the idea of Spatial Transformer Networks to introduce a novel semantic segmentation neural network. This approach decomposes the IPM process into multiple consecutive differentiable homography transform layers,  which are called "Perspective Transformer Layers". Furthermore, the interpolated feature map is refined by subsequent convolutional layers thus reducing the artifacts and improving the accuracy. The effectiveness of the proposed method in lane marking detection is validated on two public datasets: TuSimple and ApolloScape.
\end{abstract}

\section{INTRODUCTION}

Lane and road markings are critical elements in traffic scenes. The lane lines or road signs such as arrows can provide valuable information for self-driving cars' planning and controlling.

\begin{figure}[tbp]
    \centering
    \includegraphics[width=\linewidth]{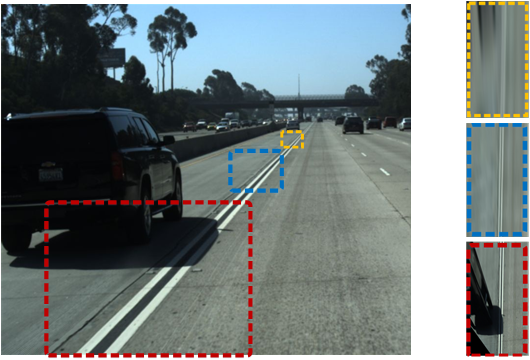}
    \caption{View of the lane markings at different distances. Similar lane markings in bird's eye view show very different shape and scale features in original view.} 
    \label{fig:lanemarking_illustration} 
    \end{figure}

Current lane marking detection methods mostly utilize the segmentation technique~\cite{VPGNet}\cite{LaneNet}\cite{SpatialCNN}, which is based on fully convolutional deep neural networks (FCNs).
The segmentation networks rely on local features which are extracted from the raw RGB pattern and mapped into semantic spaces for pixel-level classification. However, such an architecture often suffers from an accuracy degradation for lane and road markings far away from the ego-vehicle, because these markings occupy a small number of pixels in the image, and their features become inconsistent for varying distances and perspectives, as shown in Fig.~\ref{fig:lanemarking_illustration}. 
, while both distant and close lane marking information is important for the controlling and planning tasks.

\begin{figure*}[tbp]
    \centering
    \includegraphics[width=\textwidth]{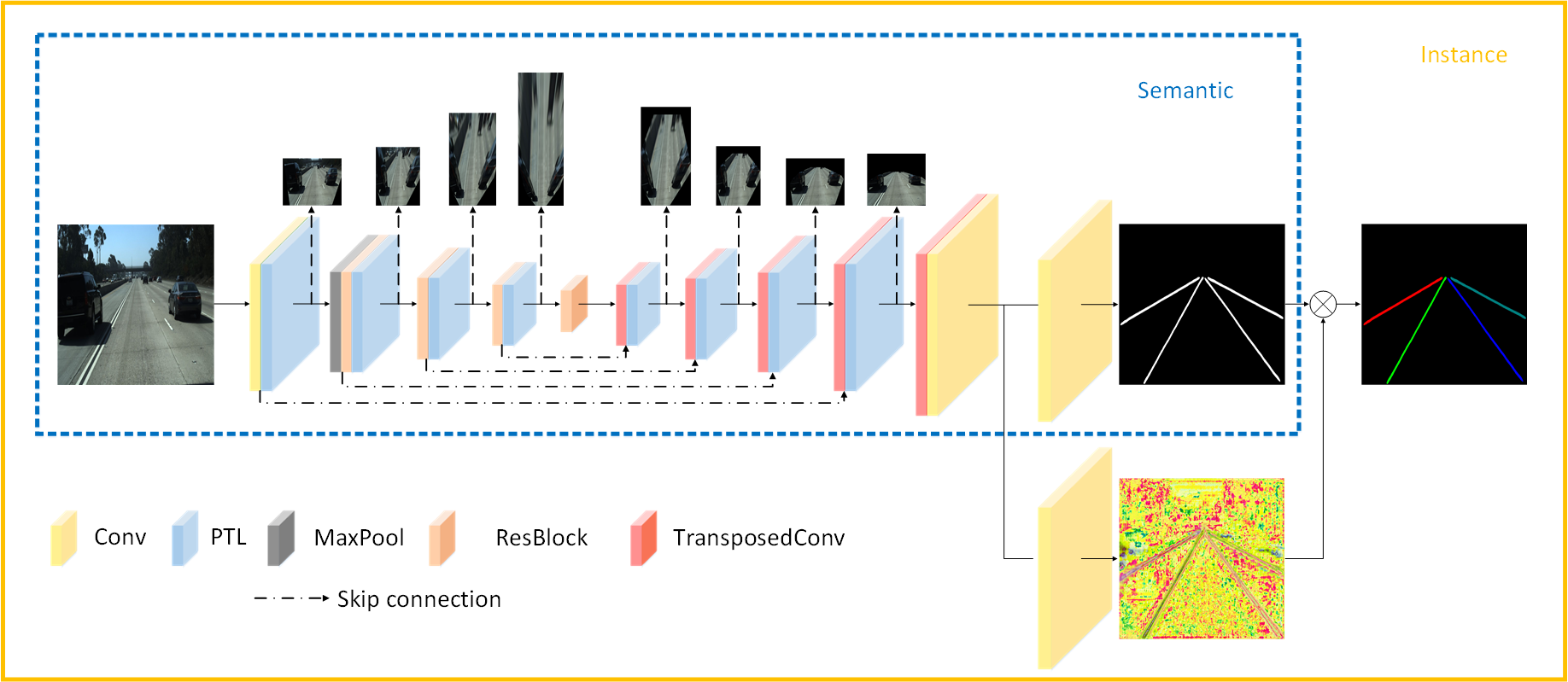}
    \caption{PTSeg architecture. Semantic segmentation network is the main body of PTSeg. It is based on a standard
    encoder-decoder network introduced in \cite{FCN}, of which the encoder is implemented with a ResNet-34 network~\cite{ResNet}. Each Perspective Transformer Layer (PTL) follows a ResBlock or a Transposed Convolution layer, perspectively and gradually warping feature maps into a bird's-eye view or back to front-view. The process of perspective transform is visualized qualitatively using color images above the main network. As for instance segmentation, following~\cite{LaneNet}, an instance embedding branch is added. It shares previous layers with the semantic segmentation networks and outputs N-dimensional embedding per lane pixel, which is also visualized as a color map.}  
    \label{fig:network_structure} 
\end{figure*}

An intuitive solution is to transform the original image to a bird's-eye view (BEV) using the Inverse Perspective Mapping (IPM). In principle, this can solve the problem of inconsistent scales of road markings at different distances. However, the IPM is typically implemented by interpolation which reduces the resolution of the distant road surface and create unnatural blurring and stretching (Fig.~\ref{fig:lanemarking_illustration}). To tackle this issue, we adopt the Encoder-Decoder architecture in Fully Convolutional Networks~\cite{FCN} and leverage the idea of Spatial Transformer Networks~\cite{STN} to build a semantic segmentation neural network.
As shown in Fig.~\ref{fig:network_structure}, Fully Convolutional layers are interleaved with a series of differentiable homographic transform layers called "Perspective Transformer Layers" (PTLs) which can transform the feature maps from the original view to the bird's-eye view during the encoding process. Afterwards, it transforms feature maps back to the original perspective in the decoding process, where subsequent convolutional layers are employed to refine each interpolated feature map.
Therefore, this network can still use the labels in the original view for an end-to-end training.

In this work, our contributions can be summarized as follows:

\begin{itemize}
    \item We proposed a lane marking detection network based on FCN, which integrates with novel PTLs to reduce the perspective distortion at a distance. 
    \item We build a mathematical model to  derive the parameters of consecutive PTLs, enabling the mutual conversion between the original view and the bird's eye view step-wisely.
    \item The effectiveness of the proposed method in both instance segmentation and semantic segmentation for lane and road markings is approved on two public datasets: TuSimple~\cite{TuSimple} and ApolloScape~\cite{apolloscape_arXiv_2018}. 
\end{itemize}{}

\section{RELATED WORKS\label{sec:related_works}}
Lane marking detection has been intensively explored and recent progresses have mainly focused on the semantic segmentation-based and instance segmentation-based methods. 
\subsection{Lane Marking Detection by Semantic Segmentation}
The work~\cite{TsinghuaRoadMarking} proposed both a road marking dataset and a segmentation network using the ResNet with Pyramid Pooling.
Lee \etal~\cite{VPGNet} proposed a unified end-to-end trainable multi-task network that jointly handles lane markings detection and road markings recognition under adverse weather conditions with the guidance by a vanishing point.
Zhang \etal~\cite{Zhang2019RoadMS} proposed a segmentation-by-detection method for road marking extraction, which delivers outstanding performances on cross datasets.
In this method, a lightweight network is dedicatedly designed for road marking detection. However, the segmentation is mainly based on conventional image morphological algorithms.

\subsection{Lane Marking Detection by Instance Segmentation}
Semantic segmentation is essentially just a pixel-level classification problem, it neither can distinguish different instances within the same category, nor can interpret separated parts of the same marking (dashed lines, zebra lines, \etc) as a unity. Therefore, researchers' attention gradually shifted to study the problem of instance segmentation.

Pan \etal~\cite{SpatialCNN} proposed the Spatial CNN (SCNN), which generalizes traditional spatial convolutions to slice-wise convolutions within feature maps, thus enabling message passing between pixels across rows and columns in a layer. This is particularly suitable for linear shaped traffic lanes. 
Hsu \etal~\cite{Proposal-Free} proposed a novel learning objective function to train the deep neural network to perform an end-to-end image pixel clustering and applied this approach on instance segmentation.
Neven \etal~\cite{LaneNet} went beyond the modeling limitation by pre-defined number of lanes, in their instance segmentation method each lane forms its own instance and the network can be trained end-to-end. They further proposed H-Net to parameterize the segmented results.

\subsection{Perspective Transform in CNNs}
To compensate the perspective distortion, a spatial transform, such as IPM, should be involved in the networks.
A typical work is the Spatial Transformer Network~\cite{STN}. It introduces a learnable module, which explicitly allows the spatial manipulation of data within the network. This differentiable module can be applied to existing convolutional architectures, enabling actively  spatial transform of feature maps.

The most similar work to ours is  \cite{Bruls2019BoostedIPM}. In this work, an adversarial learning approach is proposed for generating an improved IPM using the STN. The generated BEV images contain sharper features than that produced by traditional IPM. 
The main difference between this work and ours is that they took a ground-truth BEV image (obtained by a visual odometry) for supervision. Their target is to generate a high-resolution IPM, while ours is to improve the segmentation accuracy.

\section{PROPOSED METHOD\label{sec:method}}
In this work, we boost the performance of lane marking detection by inserting differentiable PTLs into the standard encoder-decoder architecture. One challenge in designing Transformer layers lies in dividing and distributing the integral transform into several even steps. Another is about how to determine the proper cropping range for these intermediate views. 
In this section, we firstly describe the improved backbone in section~\ref{sec:network_structure}.
Then, we address how to apply Transformer layers as well as how to solve above difficulties in section~\ref{sec:consecutive_perspective_mapping}.
Finally, we illustrate the deployment of the backbone in both semantic and instance segmentation context with details about detection heads and loss functions in~\ref{sec:loss function}. 

\subsection{Network Structure\label{sec:network_structure}}
As shown in Fig.~\ref{fig:network_structure}, the overall semantic segmentation network is based on a standard encoder-decoder network~\cite{FCN}, in which the encoder is implemented with a ResNet-34 network~\cite{ResNet} and PTLs interleave with the convolutional layers. We refer our network as PTSeg (Perspective Transformer Segmentors). In this network, images are down-sampled by the encoder to a feature map with 5 times of the stride-2 down-sampling operation. And the feature map is gradually warped into a pseudo BEV. Afterwards, the decoder reverts the previous transforms by up-sampling and back-projecting the feature map into its original size and perspective, while keeping the accumulated high-level semantic information of lane and road markings. The process of perspective transform is visualized qualitatively using color images above the main network.

As the sampling procedure does not affect the overall differentiability, the network with PTL can be trained end-to-end. Feature maps in the middle section of the network have perspective transform relationships with the input image, and the effect of these transforms in encoding process is equivalent to warping the front-view feature maps to a BEV, thus solving the problem of inconsistent scales of lane and road markings due to different distances.
Meanwhile, the subsequent refinement reduces blur and artifacts caused by interpolation. 
Similar to the FCN, skip-connections can compensate for the information loss during the down-sampling, resulting in clear boundaries for detected lane and road markings.

\subsection{Consecutive Perspective Mapping\label{sec:consecutive_perspective_mapping}}
In order to map the front-view image into a bird's-eye view smoothly, we adopt an approach differing from the standard IPM method. Here we decompose the integral transform $\mathbf{H}$ into a series of shortest-path consecutive transforms $\left\{\mathbf{H}_{i,i+1}\right\}$ (${\mathbf{H}_i}$ for short) that project the view $i$ into view $i+1$. This procedure is interpreted as
\begin{equation}\label{eqn:H_decomp_with_t_d}
\mathbf{H} = \prod_{i=0}^{N} \mathbf{H}_i \sim \prod_{i=0}^{N} \left[ \mathbf{K}_{i+1} \left(\mathbf{R}_{i,i+1}-\frac{\mathbf{t}_i\mathbf{n}_i^\top}{d_i}\right) \mathbf{K}_{i}^{-1} \right],
\end{equation}
where $\mathbf{R}_{i,i+1}$ (can be denoted as ${\mathbf{R}_i}$ for short) is the rotation matrix by which virtual camera $i+1$ is rotated in relation to virtual camera $i$; $\mathbf{t}_i$ is the translation vector from $i$ to $i+1$; $\mathbf{n}_i^\top$ and $d_i$ are the normal vector of the ground plane and the distance to the plane respectively. $\mathbf{K}_i$ and $\mathbf{K}_{i+1}$ are the cameras' intrinsic parameter matrices. 

However, to control the transform process, the value of internal parameters, \ie, $\mathbf{K}_i, \mathbf{R}_i, \mathbf{t}_i, $ and $d_i$ should be selected for each $\mathbf{H}_{i,i+1}$ by trial and error, which is a tedious job. To simplify this process, we use a pure rotation virtual camera model to eliminate $\mathbf{t}_i$, $d_i$, and use a Key-Point Bounding-Box Trick to estimate $\mathbf{K}_i$ for optimal viewports of intermediate feature maps (as shown in Algorithm~\ref{alg:optimal_view_ports}). Whereas the traditional IPM uses at least 4 pairs of pre-calibrated correspondences on each view to estimate the integral $\mathbf{H}$ directly, we estimate the integral rotation by the horizon line specified on the image, which can be obtained by horizon line detection models, \eg, the HLW\cite{HLW}. By representing the rotation in the axis-angle form, it is much easier to divide the rotation $\mathbf{R}\in \mathbb{SO}_3$ into sections by dividing the angle $\|\mathbf{\omega}\| \in \mathbb{R}$ and keep the axis direction unchanged. In this way, all internal parameters of each $\mathbf{H}_{i,i+1}$ are determined. Details about above procedure are given as follows.

\subsubsection{Pure Rotation Virtual Cameras}\, It can be proven that a translated camera with unchanged intrinsic matrix can produce the same image as a fixed camera with accordingly modified intrinsic matrix. Thus, the consecutive perspective transform is modeled as synthesizing the ground plane image captured by a pure rotating camera, and (\ref{eqn:H_decomp_with_t_d}) is simplified as,

\begin{equation}\label{eqn:H_decomp}
    \mathbf{H}_i = \mathbf{K}_{i+1} \mathbf{R}_{i,i+1} \mathbf{K}_{i}^{-1},
\end{equation}
and only the rotation matrix should be decomposed as
\begin{equation}
    \mathbf{R} = \prod_{i=0}^{N} \mathbf{R}_{i,i+1}.
\end{equation}

\subsubsection{Estimating Integral Extrinsic Rotation by the Horizon Line}
As extrinsic matrices with respect to the ground plane are not provided in TuSimple and ApolloScape datasets, we roughly estimate the integral rotation by the horizon line. Given two horizon points in the camera coordinates, $\mathbf{p}_{\text{left}}$ and $\mathbf{p}_{\text{right}}$, the normal vector of ground plane (facing to the ground) is calculated by a cross-production, \ie, 
\begin{equation}
\mathbf{n}=\frac{\mathbf{p}_{\text{left}}\times\mathbf{p}_{\text{right}}}{\|\mathbf{p}_{\text{left}}\times\mathbf{p}_{\text{right}}\|}.
\end{equation}
In order to rotate the camera to face to the ground, its $z$-axis should be rotated to align with the normal vector $\mathbf{n}$. Hence, the rotation in axis-angle form is calculated as
\begin{equation}
\mathbf{\omega} = \frac{\mathbf{e}_3 \times \mathbf{n} }{\|\mathbf{e}_3 \times \mathbf{n}\|} \cdot \text{atan2}\left(\|\mathbf{e}_3 \times \mathbf{n}\|,\mathbf{e}_3 \cdot \mathbf{n}\right),
\end{equation}
where $\mathbf{e}_3 = (0, 0, 1)^\top$ is a unit vector on $z$-axis. 

\subsubsection{Decomposing the Extrinsic Rotation}
Here we use the axis-angle representation for decomposing the rotation. We simply divide the integral angle into several even parts, and then convert each $\mathbf{\omega}_i$ to the corresponding rotation matrix $\mathbf{R}_i$.

\subsubsection{Optimal Viewports by Key-Point Bounding Boxes}
While conducting IPM, image pixels at the edge often need to be cropped to prevent the target view from being too large. In order to preserve the informative pixels as many as possible, we roughly annotate the ground region by a set of border points in the front-view. The points are projected to the new view during each perspective transform. And we use a bounding box in the new view to determine the minimal available viewport which does not crop any projected key point. Thus, given a desirable target view width $W_{i+1}$, the corresponding intrinsic $K_{i+1}$ and target view height $H_{i+1}$ is determined, as shown in Algorithm \ref{alg:optimal_view_ports}.

\begin{algorithm}[tbp]
    \caption{Determine the Optimal Viewports through Key-Point Bounding Boxes}\label{alg:optimal_view_ports}
    \begin{algorithmic}[1]
        \REQUIRE $\mathbf{K}_i$, $\mathbf{R}_i$, $\mathbf{p}_{i,j}^I$, $W_{i+1}$
        \STATE Convert points from Image $I$ to Camera $C$: $\mathbf{p}_{i,j}^C = \mathbf{K}_i^{-1} \mathbf{p}_{i,j}^I$
        \STATE Rotate points to view $i+1$: $\mathbf{p}_{i+1,j}^C = \mathbf{R}_i \mathbf{p}_{i,j}^C$
        \STATE Normalize by the Z-dimension: $\mathbf{p}_{i+1,j}^C /= \text{abs}\left(Z_{\mathbf{p}_{i+1,j}^C}\right)$
        \STATE Get the bounding box: [$bb_{top}, bb_{left}, bb_{width}, bb_{height}] = bbox(\mathbf{p}_{i+1,:}^C)$
        \STATE Estimate focal length $f_{i+1}$ as a scale ratio: $f_{i+1} = W_{i+1} / bb_{width}$
        \STATE Estimate target view height with the same scale ratio: $H_{i+1} = f_{i+1} \times bb_{height}$
        \STATE Estimate translation $t_{i+1}$ by aligning left-top corner of target image view and bounding box in target camera coordinate: $t_{i+1} = {(0, 0)}^\top - {(f_{i+1} \times bb_{left}, f_{i+1}\times bb_{top})}^\top$
        \STATE Compose target intrinsic matrix: $ \mathbf{K}_{i+1} = (f_{i+1}, t_{i+1}) $
        \RETURN $\mathbf{K}_{i+1}$, $H_{i+1}$
    \end{algorithmic}
\end{algorithm}

\subsection{Segmentation Heads\label{sec:loss function}}

\subsubsection{Semantic Segmentation\label{sec:semantic_segmentation_head}}

The lane and road marking detection problems are often cast as a semantic segmentation task~\cite{VPGNet}~\cite{TsinghuaRoadMarking}~\cite{Zhang2019RoadMS}. By representing label classes as one-hot vectors, we predict the logits of each class at each pixel location. Then, we use the classic cross-entropy loss function to train this semantic segmentation branch.

\subsubsection{Instance Segmentation\label{sec:instance_segmentation_head}}

We follow the work of LaneNet~\cite{LaneNet} to interpret the lane detection problem as an instance segmentation task. The network contains two branches. The semantic branch outputs a binary mask, while the instance embedding branch outputs an N-dimensional embedding vector for each pixel. In the embedding space pixels are more easily to be clustered by a one-shot method based on distance metric learning~\cite{brab2017semantic}. For details of the loss function please refer to~\cite{LaneNet}.

\section{EXPERIMENTS\label{sec:experiments}}
We evaluate our network on the TuSimple~\cite{TuSimple} and ApolloScape~\cite{apolloscape_arXiv_2018} dataset respectively for the instance segmentation and semantic segmentation tasks. Our network is implemented by the PyTorch~\cite{PyTorch} framework.

\subsection{TuSimple Benchmark}
\subsubsection{Dataset}
The TuSimple Benchmark is a dedicated dataset for lane detection and consists of 3626 training and 2782 testing images. The annotation includes the $x$-position of the lane points at a number of discretized $y$-positions.

\subsubsection{Metrics}
The detection accuracy is calculated as the average correct number of points per image:
\begin{equation}
    \mathrm{acc}=\sum_{i m} \frac{C_{i m}}{S_{i m}},
\end{equation}
where $C_{i m}$ denotes the number of correct points and $S_{i m}$ is the number of groundtruth points. A point is regarded as correctly detected when the error is smaller than a predefined threshold. Besides, the false positive and false negative scores can also be calculated by
\begin{equation}
\mathrm{FP} =\frac{F_{\text {pred}}}{N_{\text {pred}}} , ~~~
\mathrm{FN} =\frac{M_{\text {pred}}}{N_{g t}},
\end{equation}
where $F_{\text {pred}}$ denotes the number of mispredicted lanes, $N_{\text {pred}}$ indicates the number of predicted lanes, $M_{\text {pred}}$ is the number of missed groundtruth lanes and $N_{\text {gt}}$ represents the number of all groundtruth lanes.

\subsubsection{Training Details}
Here we train the instance segmentation network as shown in Fig.~\ref{fig:network_structure}. During the training process we use the Adam~\cite{kingma2014adam} optimizer, with a weight decay of 0.0005, a momentum of 0.95, a learning rate of 0.00004, and a batch size of 2. When the accuracy is without promotion up to 60 epochs, the learning rate drops to 10\%. And the model converges after 220 epochs.

 \begin{table}[tbp]
\caption{Test results on TuSimple Lane Detection Benchmark. \hspace{\textwidth} \footnotesize{ (u.h. is short for the region under the horizon line.)}}
\label{tab:tusimple_compare_with_others}
	\resizebox{0.98\linewidth}{!}{ %
		\setlength\tabcolsep{5pt}
\begin{tabular}{llllll}
\hline
                       & ACC   & FP     & FN     & Ext.data   & ACC(u.h.)\\ \hline
Xingang Pan{~\cite{SpatialCNN}}      & 96.53 & 0.0617 & 0.018  & yes  & N/A      \\
Yen-Chang Hsu{~\cite{Proposal-Free}}    & 96.50  & 0.0851 & 0.0269 & no  & N/A       \\
Davy Neven{~\cite{LaneNet}}       & 96.40  & 0.078  & 0.0244 & no  & N/A       \\ \hline
ResNet34-FCN     & 96.24 & 0.0746 & 0.0347 & no  & 95.67       \\
ResNet34-PTL-FCN (ours) & 96.15 & 0.0818 & 0.0314 & no  & 95.72       \\ \hline
\end{tabular}
}
\end{table}

\begin{figure}[tp]
    \centering
    \vspace*{-1.0cm}
    \subfigure[Lane points accuracy vs. distance in pixels.]{
        \includegraphics[width=\linewidth, trim = 0mm 0mm 0mm 12mm, clip=true]{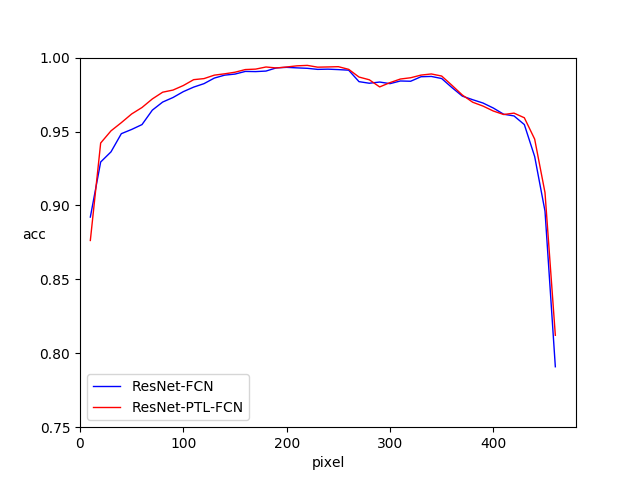}
    }%
    \\
    \subfigure[Lane points accuracy vs. distance in meters.]{
        \includegraphics[width=\linewidth, trim = 0mm 0mm 0mm 12mm, clip=true]{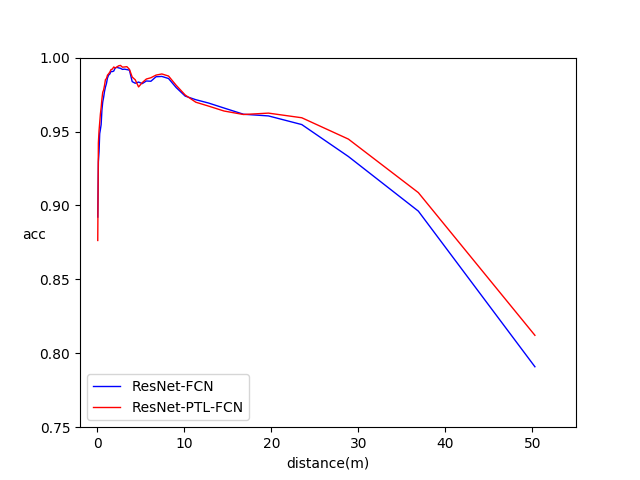}
    }%
    \caption{TuSimple Lane Detection Benchmark Results.}
    \label{fig:compare_tusimple}
\end{figure}

\begin{figure}[tp]
    \centering
    \includegraphics[width=\linewidth]{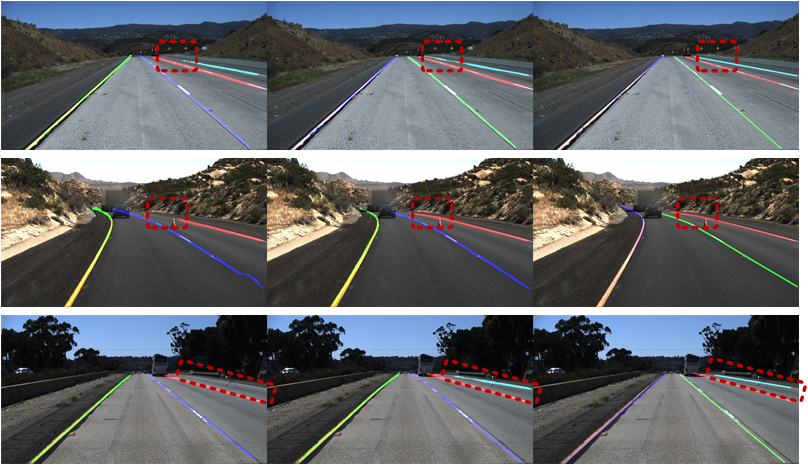}
    \caption{Visualization of the comparison among the base-line, our method and the groundtruth on Tusimple dataset. Each row contains three submaps. From left to right: results w/o PTLs, results w/ PTLs, GT. And the area inside the red wireframe should be paid more attention. }
    \label{fig:Visua_Tusimple} 
    \end{figure}
    
\subsubsection{Evaluation Results}
In comparison with other state-of-the-art methods~\cite{LaneNet}~\cite{SpatialCNN}~\cite{Proposal-Free}, we show the test results in Table~\ref{tab:tusimple_compare_with_others}, from which we can find out that our detection accuracy is already in the first echelon. 
It is worth to mention that all evaluation results above are in strict accordance with the metric defined by TuSimple. However, in our method the feature maps are warped into a bird's-eye view of the ground, which force a part of the image above the horizon to be ignored by our method. An amount of lane segments existing above the horizon are visible for other methods but invisible for our method, which would lead to a slight decrease of the results. In order to make a fair comparison, we re-evaluated those samples only below the horizon, and ignored null sample points which is labeled as $-2$ in the annotation\footnote{See the labeling protocol in TuSimple.}. The new evaluation result is listed in $ACC(under horizon)$ column of Table~\ref{tab:tusimple_compare_with_others}. We also plot the accuracy versus different distances from ego-vehicle in the line charts.
Fig.~\ref{fig:compare_tusimple}~(a) shows the accuracy in dependence of pixel distance to the image bottom. Fig.~\ref{fig:compare_tusimple}~(b) shows the accuracy in dependence of the real distance to the ego-vehicle. Both charts imply that our method can improve the detection accuracy of lane and road markings at longer distances.
The qualitative comparison is shown in Fig.~\ref{fig:Visua_Tusimple}.

\subsection{ApolloScape Benchmark}
\subsubsection{Dataset}
The Lane Segmentation branch in ApolloScape dataset contains more than 110,000 frames with high quality pixel-level annotations. The annotation includes 35 kinds of lane and road markings from daily traffic scenarios, including but not limited to lanes, turning arrows, stop lines, zebra crossings. 
To the best of the authors' knowledge, no related works have been trained on the ApolloScape Lane Segmentation dataset. Therefore, we only show the ablation experimental results of our own method.

\subsubsection{Metrics}
The evaluation follows the recommendation of ApolloScape which uses the mean-IOU (mIOU) as the evaluation metric just like in~\cite{Cordts2016Cityscapes}.

\begin{figure}[tbp]
    \centering
    \vspace*{0.0cm}
    \subfigure[mean-IOU vs. distance in pixels]{
        \includegraphics[width=\linewidth, trim = 0mm 0mm 0mm 12mm, clip=true]{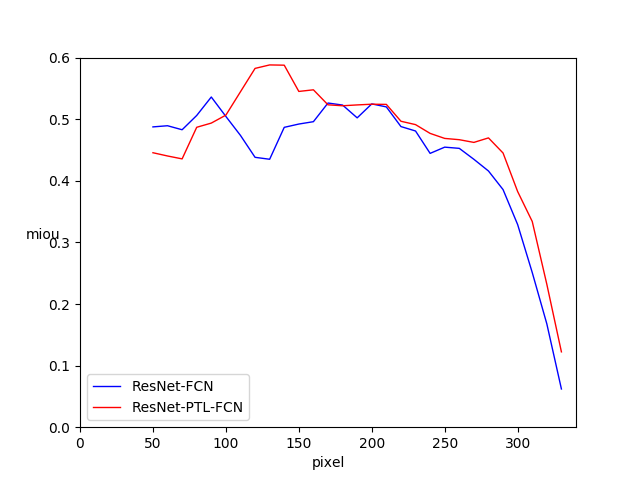}
    }%
    \\
    \subfigure[mean-IOU vs. distance in meters]{
        \includegraphics[width=\linewidth, trim = 0mm 0mm 0mm 12mm, clip=true]{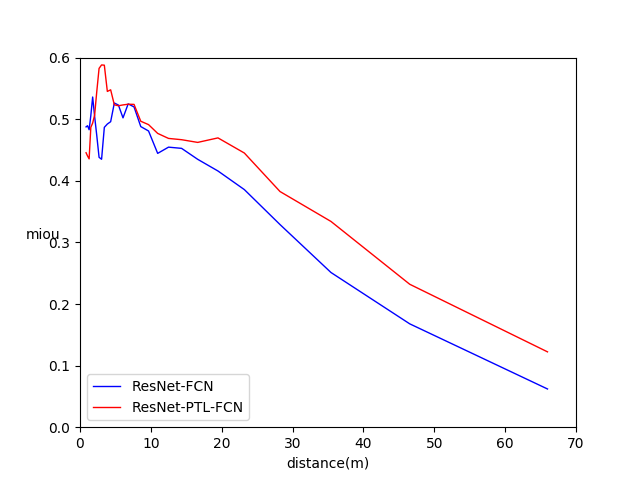}
    }%
    \caption{Apollo Road Marking Semantic Segmentation Results.}
    \label{fig:compare_apollo}
\end{figure}

\subsubsection{Training Details}
Since the ApolloScape only provides pixel-level semantic annotations, we train the semantic segmentation network as shown in Fig.~\ref{fig:network_structure}.
During the training process we use the Adam~\cite{kingma2014adam} optimizer, with a weight decay of 0.0005, a momentum of 0.95, a learning rate of 0.00004, and a batch size of 2. And the model converges after 25 epochs.

\subsubsection{Evaluation Results}
Also in a fair way, we evaluate the image part below the horizon. Fig.~\ref{fig:compare_apollo} shows the results of mean-IOU accuracy at different distances. 
Table~\ref{tab:apollo_per_class_iou} shows mIOU value and IOU values of some common types of lane and road markings. We ignored the rest classes who accounting for less than 0.1\% of the data.

\begin{table}[tbp]
\caption{Per-class IOU results on ApolloScapes Lane Segmentation Benchmark.}
\label{tab:apollo_per_class_iou}
	\resizebox{0.98\linewidth}{!}{ %
		\setlength\tabcolsep{5pt}
\begin{tabular}{llcc}
\hline
category                   & class               & ResNet18-FCN      & ResNet18-PTL-FCN (ours)  \\ \hline
\multirow{4}{*}{arrow} & thru                & 0.611          & \textbf{0.692} \\
                       & thru \& left turn   & 0.768          & \textbf{0.800}  \\
                       & thru \& right turn  & \textbf{0.824} & 0.808          \\
                       & left turn           & 0.767          & \textbf{0.768} \\ \hline
stopping               & stop line           & 0.665          & \textbf{0.747} \\ \hline
zebra                  & crosswalk           & \textbf{0.859}  & 0.858           \\ \hline
\multirow{4}{*}{lane}  & white solid         & \textbf{0.832} & 0.800          \\
                       & yellow solid        & \textbf{0.813} & 0.803          \\
                       & yellow double solid & 0.886          & \textbf{0.893} \\
                       & white broken        & \textbf{0.791} & 0.790          \\ \hline
diamond                & zebra attention     & 0.749          & \textbf{0.775} \\ \hline
rectangle              & no parking          & 0.652           & \textbf{0.724} \\ \hline
mIOU              &            & 0.768           & \textbf{0.788} \\ \hline
\end{tabular}
}
\end{table}

\begin{figure}[tbp]
    \centering
    \includegraphics[width=\linewidth]{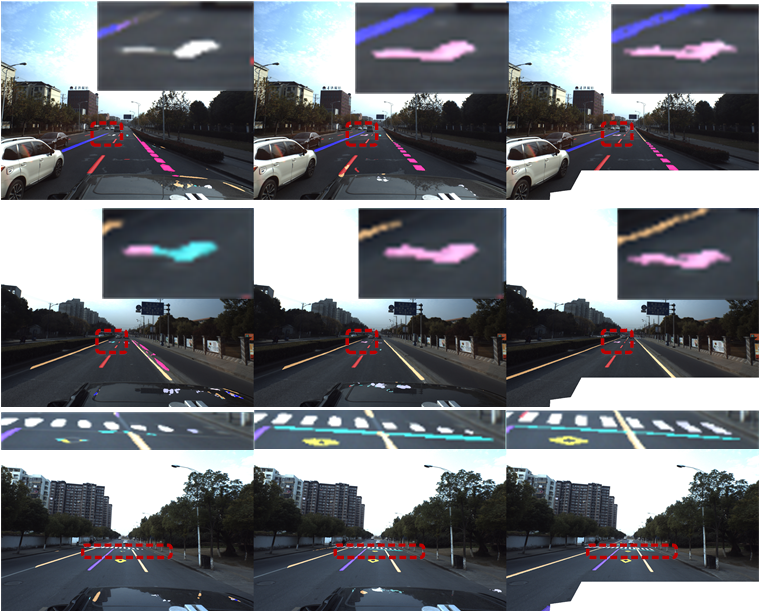}
    \caption{Visualization of the comparison among the base-line, our method and the ground-truth on ApolloScape. Each row contains three submaps. From left to right: results w/o PTLs, results w/ PTLs, GT, and the enlarged view of area inside the red wireframe is placed on the top of each submap.} 
    \label{fig:Visua_Apollo} 
    \end{figure}

According to the experimental results, our method can effectively improve the detection accuracy at further distances, especially for the road markings with richer structural features such as turning arrows.
The qualitative comparison is shown in Fig.~\ref{fig:Visua_Apollo}. 

\section{CONCLUSION\label{sec:conclusion}}
In this paper, we introduced a segmentation network architecture improved by consecutive homography transforms for road marking detection. The parameters of consecutive transforms are clearly yielded by a pure rotating camera model and a key-point bounding-box trick. The proposed method is proven to be beneficial for distant lane and road marking detection.
For the future research, we are going to incorporate an online scheme of extrinsic estimation into this structure. Also handling the non-flat ground surface which has unclear definition of ground normal vectors and horizon lines is one of the interesting topics. 

\section*{Acknowledgment}
This work is supported by the National Key Research and Development Program of China (No. 2018YFB0105103, No. 2017YFA0603104), the National Natural Science Foundation of China (No. U1764261, No. 41801335, No. 41871370), the Natural Science Foundation of Shanghai (No. kz170020173571, No. 16DZ1100701) and the Fundamental Research Funds for the Central Universities (No. 22120180095).


\bibliographystyle{unsrt}
\bibliography{wpref}

\begin{thebibliography}{10}

\bibitem{VPGNet}
Seokju Lee, Junsik Kim, Jae~Shin Yoon, Seunghak Shin, Oleksandr Bailo, Namil
  Kim, Tae-Hee Lee, Hyun~Seok Hong, Seung-Hoon Han, and In~So Kweon.
\newblock Vpgnet: Vanishing point guided network for lane and road marking
  detection and recognition.
\newblock {\em arXiv:1710.06288 [cs]}, Oct 2017.
\newblock arXiv: 1710.06288.

\bibitem{LaneNet}
Davy Neven, Bert De~Brabandere, Stamatios Georgoulis, Marc Proesmans, and Luc
  Van~Gool.
\newblock Towards end-to-end lane detection: an instance segmentation approach.
\newblock {\em arXiv:1802.05591 [cs]}, Feb 2018.
\newblock arXiv: 1802.05591.

\bibitem{SpatialCNN}
Xingang Pan, Jianping Shi, Ping Luo, Xiaogang Wang, and Xiaoou Tang.
\newblock Spatial as deep: Spatial cnn for traffic scene understanding.
\newblock 2018.

\bibitem{FCN}
Jonathan Long, Evan Shelhamer, and Trevor Darrell.
\newblock Fully convolutional networks for semantic segmentation.
\newblock In {\em Proceedings of the IEEE conference on computer vision and
  pattern recognition}, pages 3431--3440, 2015.

\bibitem{STN}
Max Jaderberg, Karen Simonyan, and Andrew Zisserman.
\newblock Spatial transformer networks.
\newblock page~9.

\bibitem{TuSimple}
Tusimple lane detection benchmark.
\newblock \url{http://benchmark.tusimple.ai/}.

\bibitem{apolloscape_arXiv_2018}
Xinyu Huang, Xinjing Cheng, Qichuan Geng, Binbin Cao, Dingfu Zhou, Peng Wang,
  Yuanqing Lin, and Ruigang Yang.
\newblock The apolloscape dataset for autonomous driving.
\newblock {\em arXiv: 1803.06184}, 2018.

\bibitem{ResNet}
Kaiming He, Xiangyu Zhang, Shaoqing Ren, and Jian Sun.
\newblock Deep residual learning for image recognition.
\newblock {\em 2016 IEEE Conference on Computer Vision and Pattern Recognition
  (CVPR)}, Jun 2016.

\bibitem{TsinghuaRoadMarking}
Xiaolong Liu, Zhidong Deng, Hongchao Lu, and Lele Cao.
\newblock Benchmark for road marking detection: Dataset specification and
  performance baseline.
\newblock In {\em 2017 IEEE 20th International Conference on Intelligent
  Transportation Systems (ITSC)}, pages 1--6. IEEE, 2017.

\bibitem{Taoyang2018}
Y.~{Wu}, T.~{Yang}, J.~{Zhao}, L.~{Guan}, and W.~{Jiang}.
\newblock Vh-hfcn based parking slot and lane markings segmentation on
  panoramic surround view.
\newblock In {\em 2018 IEEE Intelligent Vehicles Symposium (IV)}, pages
  1767--1772, June 2018.

\bibitem{Zhang2019RoadMS}
Weiwei Zhang, Zeyang Mi, Yaocheng Zheng, Qiaoming Gao, and Wenjing Li.
\newblock Road marking segmentation based on siamese attention module and
  maximum stable external region.
\newblock {\em IEEE Access}, 7:143710--143720, 2019.

\bibitem{Proposal-Free}
Yen-Chang Hsu, Zheng Xu, Zsolt Kira, and Jiawei Huang.
\newblock Learning to cluster for proposal-free instance segmentation.
\newblock {\em arXiv:1803.06459 [cs]}, Mar 2018.
\newblock arXiv: 1803.06459.

\bibitem{Bruls2019BoostedIPM}
Tom Bruls, Horia Porav, Lars Kunze, and Paul Newman.
\newblock The right (angled) perspective: Improving the understanding of road
  scenes using boosted inverse perspective mapping.
\newblock In {\em 2019 IEEE Intelligent Vehicles Symposium (IV)}, page
  302–309, Jun 2019.

\bibitem{HLW}
Scott Workman, Menghua Zhai, and Nathan Jacobs.
\newblock Horizon lines in the wild.
\newblock In {\em Procedings of the British Machine Vision Conference 2016},
  pages 20.1--20.12. British Machine Vision Association, 2016.

\bibitem{brab2017semantic}
Bert~De Brabandere, Davy Neven, and Luc~Van Gool.
\newblock Semantic instance segmentation with a discriminative loss function,
  2017.

\bibitem{PyTorch}
Adam Paszke, Sam Gross, Soumith Chintala, Gregory Chanan, Edward Yang, Zachary
  DeVito, Zeming Lin, Alban Desmaison, Luca Antiga, and Adam Lerer.
\newblock Automatic differentiation in pytorch.
\newblock 2017.

\bibitem{paszke2016enet}
Adam Paszke, Abhishek Chaurasia, Sangpil Kim, and Eugenio Culurciello.
\newblock Enet: A deep neural network architecture for real-time semantic
  segmentation, 2016.

\bibitem{kingma2014adam}
Diederik~P Kingma and Jimmy Ba.
\newblock Adam: A method for stochastic optimization.
\newblock {\em arXiv preprint arXiv:1412.6980}, 2014.

\bibitem{Cordts2016Cityscapes}
Marius Cordts, Mohamed Omran, Sebastian Ramos, Timo Rehfeld, Markus Enzweiler,
  Rodrigo Benenson, Uwe Franke, Stefan Roth, and Bernt Schiele.
\newblock The cityscapes dataset for semantic urban scene understanding.
\newblock In {\em Proc. of the IEEE Conference on Computer Vision and Pattern
  Recognition (CVPR)}, 2016.

\end{thebibliography}


\end{document}